\documentclass[review]{elsarticle}

\usepackage{lineno,hyperref}
\usepackage[utf8]{inputenc}
\usepackage{amsmath,amssymb}
\usepackage{natbib}
\usepackage{color}
\usepackage{url}
\usepackage{caption}
\usepackage{subcaption}
\usepackage{fancyvrb}
\usepackage{listings}
\usepackage{hhline}
\usepackage{float}
\usepackage{adjustbox}
\usepackage{amsmath}
\usepackage{nonfloat}
\usepackage{multirow}
\usepackage[table,xcdraw]{xcolor}
\usepackage{todonotes}
\usepackage{xcolor}
\usepackage{algorithm}
\usepackage{algpseudocode}
\usepackage{footmisc}

\journal{Journal of \LaTeX\ Templates}

\begin{document}

\begin{frontmatter}

%\title{Bilingual Embeddings\\ via Constraints and Random Walks over WordNet}
\title{Bilingual Embeddings\\ with Random Walks over Multilingual Wordnets}

%% Group authors per affiliation:
\author{Josu Goikoetxea\footnote[2]{Corresponding author. \textit{Email adress:} \textsf{josu.goikoetxea@ehu.eus}}, Aitor Soroa, Eneko Agirre}
\address{IXA NLP group, Faculty of Informatics, UPV/EHU, \\ Manuel Lardizabal 1 (20018), Donostia, Basque Country\vspace{-1.02cm}}

%\fntext[myfootnote]{Since 1880.}

%% or include affiliations in footnotes:
%\author[mymainaddress,mysecondaryaddress]{Josu Goikoetxea Salutregi, Eneko Agirre Bengoa, Aitor Soroa Etxabe, Mikel Artetxe Zurutuza \\ 
%        \{josu.goikoetxea,e.agirre, a.soroa, mikel.artetxe\}@ehu.eus \\
%         IXA Group}
%\ead[url]{www.elsevier.com}

%\author[mysecondaryaddress]{Global Customer Service\corref{mycorrespondingauthor}}
%\cortext[mycorrespondingauthor]{Corresponding author}
%\ead{support@elsevier.com}

%\address[mymainaddress]{1600 John F Kennedy Boulevard, Philadelphia}
%\address[mysecondaryaddress]{360 Park Avenue South, New York}

\begin{abstract}

Bilingual word embeddings represent words of two languages in the same space, and allow to transfer knowledge from one language to the other without machine translation. The main approach %, which has been shown to be effective in cross-lingual applications, 
is to train monolingual embeddings first and then map them using bilingual dictionaries. In this work, we present a novel method to learn bilingual embeddings based on multilingual knowledge bases (KB) such as WordNet. Our method extracts bilingual information from multilingual wordnets via random walks and learns a joint embedding space in one go. We further reinforce cross-lingual equivalence adding bilingual constraints in the loss function of the popular skipgram model. Our experiments involve twelve cross-lingual word similarity and relatedness datasets in six language pairs covering four languages, and show that:
  1) random walks over multilingual wordnets improve results over just using dictionaries; 2) multilingual wordnets on their own improve over text-based systems in similarity datasets; 3) the good results are consistent for large wordnets (e.g. English, Spanish), smaller wordnets (e.g. Basque) or loosely aligned wordnets (e.g. Italian); 4) the combination of wordnets and text yields the best results, above mapping-based approaches. Our method can be applied to richer KBs like DBpedia or BabelNet, and can be easily extended to multilingual embeddings. All software and resources are open source.
\end{abstract}

\begin{keyword}
\texttt multilinguality\sep distributional semantics\sep embeddings\sep random walks\sep WordNet
%\MSC[2010] 00-01\sep  99-00
\end{keyword}

\end{frontmatter}

%\linenumbers

\newcommand\tab[1][1.2cm]{\hspace*{#1}}

\section{Introduction}
\label{sec:intro}

%\todo{BEgiratu KB vs. wordnet, eta esan nahiz eta metodoa wordnet-ekin egin dugun, printzipios beste multilingual KBekin egin daitekela e.g. wikipedia, babelnet} 
The use of distributional semantic models in the Natural Language Processing community has grown substantially over the last few years \citep{mikolov2013efficient}. In particular, unsupervised distributional
models such as skipgram \citep{mikolov2013distributed} and GloVe \citep{pennington2014glove} represent semantic and syntactic features of
words in low dimensional embeddings learned using co-occurrence information from corpora. These word embeddings have been very effective when addressing a
wide variety of tasks, such as word similarity \citep{baroni2014don}, word analogy \citep{levy2014neural,mikolov2013distributed}, sentiment analysis \citep{tang2014learning}, document classification \citep{hermann-blunsom:2014:P14-1} and natural language inference \citep{bowman2015large}, to name a few. 

Several authors \cite{halawi2012large,faruqui-EtAl:2015:NAACL-HLT,rothe-schutze:2015:ACL-IJCNLP,bollegala2016joint,goikoetxea2016single} have proposed methods to combine text-based distributional information with the semantic information contained in Knowledge Bases (KB) such as WordNet\cite{miller1995wordnet}. The combination improves the results over corpus-only methods, showing that text and knowledge bases contain complementary information. Our work can be seen as an extension of those ideas to a bilingual setting.

% Knowledge Bases such as WordNet \citep{miller1995wordnet} contain relevant semantic information. Some authors have explored methods to combine them with co-occurrence information into a single representation space \citep{halawi2012large,faruqui:2015:Retro,rothe2015autoextend,bollegala2015joint,goikoetxea2016single}, improving results over corpus-only methods and showing that text and knowledge bases contain complementary information. Our work can be seen as an extension of those ideas to a bilingual setting.

Given the success of distributional models on monolingual tasks, researchers have also trained bilingual word embeddings that represent words in different languages within the same vector space. The approaches can be classified according to the bilingual resources used. Some methods require 
%which range from parallel corpora or aligned bilingual documents to bilingual dictionaries. The first one requires 
large parallel corpora \citep{gouws2015bilbowa,luong2015bilingual}, which is a scarce resource for many language pairs. Other methods use more widely available document-aligned corpora \citep{sogaard2015inverted,vulic2016bilingual,mogadala2016bilingual}.
Finally, other alternative approaches rely on bilingual dictionaries to learn a linear transformation that maps independently-learned monolingual embeddings \citep{mikolov2013linguistic,artetxe-labaka-agirre:2016:EMNLP2016,xing2015normalized}. 
The latter setting only requires monolingual corpora and a small bilingual dictionary (around 5,000 word pairs), which can be obtained easily for many language pairs. 

We chose the latter setting for this article, but, instead of bilingual dictionaries, we turn our attention to multilingual wordnets\footnote{We follow the literature using capitalized WordNet for the English version and wordnets in lowercase to refer to instantiations in other languages. As we focus on two languages, we will talk about bilingual wordnets, although they are truly multilingual.} \citep{vossen1998multilingual}. Bilingual dictionaries are basically pairs of translation-equivalent words, while bilingual wordnets also contain a wealth of information about the structure of the lexicon, including semantic relations like synonymy, hypernymy (aka is-a), meronymy (aka part-of), antonymy, and gloss-relations \citep{miller1995wordnet}. Our hypothesis is that the structural information in wordnets can be exploited to produce better bilingual embeddings, compared to methods that just use the word pairs in bilingual dictionaries. 

%In this paper, we extend the mentioned methods which combine text and KB semantic information; that is, we have adapted the techniques implemented for enriching embeddings within a single monolingual space in order to strengthen cross-lingual similarity between languages within a vector space. Implicitly, we exploit the language-independent semantic relations among concepts in KBs
The main contribution of this article is thus a novel method based on random walks over concepts in the multilingual KBs to produce bilingual embeddings: each time the random walker arrives at a concept, it emits, at random, one possible lexicalization in any of the two languages. These emitted words constitute a bilingual synthetic corpus which encodes, via bilingual co-occurrences, the latent structure of the bilingual wordnet. 
%the in words from any of the two languages mapped in WordNet, thereby creating a pseudo-corpora with cross-lingual co-occurrences which latently contains
%the WOrdNet structure.
%resides in the extension of the algorithm proposed by \citep{goikoetxea2015random} to two languages for performing bilingual random walks over WordNet. The algorithm performs random walks over language-independent concepts and emits words from any of the two languages mapped in WordNet, thereby creating a pseudo-corpora with cross-lingual co-occurrences which latently contains
%the WOrdNet structure. 
We feed the bilingual co-occurrences into the skipgram model \citep{mikolov2013efficient}, so that words in the two languages that are structurally similar in the KB have similar embeddings. A similar strategy was previously shown to be very effective for monolingual wordnets \citep{goikoetxea2015random}, and we now extend it to bilingual wordnets with excellent results. The proposed method is independent of the particular KB used, and other multilingual KBs such as Dbpedia or BabelNet \citep{navigli2012} can be considered.

%as \citep{goikoetxea2016single} authors, who successfully proved that text and KB semantic information is complementary. Contrary to the mentioned mapping methods, our approach learns from scratch and uses a single vector space. \todo{enekoren errebisioa honeraino iritsi da}
In addition, we also explore the use of bilingual constraints extracted from wordnets in the loss function of skipgram. These constraints act as regularizers, implementing the notion that two words which are translations of each other (according to the  bilingual wordnet) should have embeddings which are close in embedding space. This can be seen as an extension of \citep{halawi2012large,bollegala2016joint} to the bilingual case.  
%Also inspired by originally monolingual methods (\citep{halawi2012large}, \citep{bollegala2015joint}), and as a complement to the previous approach, we added a regularizer in the former word2vec's skipgram model to jointly learn from text co-occurrences and KB relational information. Instead of monolingual constraints we introduced bilingual ones extracted from cross-lingual equivalences in WordNet, so that we constrained the neighboring words' embeddings in a KB across languages to be similar. Further, as we will see in section \ref{sec:exp}, we combined bilingual constraints with bilingual random walks reinforcing cross-lingual similarity and thus proving the compatibility of our approaches. 
%\todo[color=blue]{contributions hobeto ekarri ona, edo abstract-ean utzi?}

In order to complement the WordNet structure with distributional information, we combine the bilingual synthetic corpus with monolingual text corpora, and then learn bilingual embeddings using skipgram in one go. This method, coupled with the bilingual constraints, improves the state-of-the-art in twelve cross-lingual similarity and relatedness datasets comprising five language pairs and four languages.

The article is structured as follows. We start presenting related work. Section \ref{sec:constraints} presents the skipgram model and the modified loss function which includes bilingual constraints. Section \ref{sec:rw} presents the random walk method for learning embeddings from bilingual knowledge bases. Section \ref{sec:exp} presents the experiments on cross-lingual word similarity and relatedness, followed by a summary and comparison to the state-of-the-art. Finally, we draw the conclusions and future work. 

%how we modify the method to learn monolingual embeddings in order to introduce bilingual constraints from bilingual dictionaries so we jointly learn bilingual embeddings in one go. Section \ref{sec:rw} presents how we extend the method to learn embeddings from monolingual knowledge bases to  bilingual knowledge bases. Section \ref{sec:exp}  presents the experiments on cross-lingual word similarity and relatedness, followed by the conclusions and future work. 

\section{Related work}
\label{sec:relwork}

Several methods to learn word embeddings from text corpora exist, being skipgram \citep{mikolov2013distributed}, continuous bag of words \citep{mikolov2013distributed} and GloVe \citep{pennington2014glove} the most popular. Without lose of generality, in this work we have chosen the most widely used, skipgram, although the techniques here presented can be easily applied to other embedding learning models. 

Following the hypothesis that text  and knowledge bases contain  complementary semantic information, several authors have tried to enhance
the quality of distributional representations by incorporating information from knowledge bases. \citep{faruqui-EtAl:2015:NAACL-HLT}
proposed a  method to infuse relational information from WordNet \citep{miller1995wordnet}, Paraphrase Database
\citep{ganitkevitch2013ppdb}, FreeBase \citep{bollacker2008freebase} and FrameNet \citep{baker1998berkeley} to embeddings learned from
corpora, improving significantly the results in word similarity tasks. 

Some other authors learn embeddings jointly using knowledge base relations and corpus co-occurrences. \citep{rastogi-EtAl:2015:NAACL-HLT}
implemented a LSA-based multiview model using several semantic resources. \citep{wang2014knowledge} introduced relational information from
FreeBase in the learning process and performed projections of entity embeddings onto an hyperplane w.r.t. the knowledge base relation. \citep{halawi2012large}
and \citep{bollegala2016joint} incorporated WordNet information introducing a regularizer in the loss function which constrained the neighbors in
the knowledge base to have similar representations. In this article we have extended the latter technique, incorporating bilingual constraints extracted
from bilingual wordnets into the skipgram loss function (see section \ref{sec:biconst}).
%Thereby, our method will compute similar
%embeddings for neighboring words in WordNet across languages within the same vector space.

In a different strand of work, \citep{goikoetxea2015random} learn knowledge-based embeddings using graph-based methods over WordNet; they performed random walks over  WordNet relations to  create a synthetic corpus that implicitly
encoded the structure of the knowledge base, similar to the DeepWalk method \citep{Perozzi:2014:DOL:2623330.2623732}. In later work, they show that the synthetic corpus can be combined with text corpora, yielding embeddings which combine both sources of information, and surpassing other methods in word similarity tasks \citep{goikoetxea2016single}.
In this work we extend the random walk algorithm to work with bilingual wordnets, both to produce knowledge-based bilingual embeddings, and hybrid text- and knowledge-based bilingual embeddings (see Section \ref{sec:rw}).
%, which, along with the cross-lingual constraints,
%reinforces the representation of equivalent words across languages.
%Along with the cross-lingual constraints and in relation with the random walks, in order to reinforce equivalent words' representations across languages,  we have extended the random walk algorithm from monolingual to bilingual.

%Furthermore, based on \citep{goikoetxea2016single} author's combination of monolingual text- and WordNet-corpus (see section \ref{sec:rw}),
%we combined the mentioned bilingual WordNet synthetic corpus with bilingual text corpus. We will explain in detail our new approaches in section
%\ref{sec:birw}.

Bilingual distributional representations have also attracted a lot of
attention, and researchers have addressed the training of bilingual
embeddings using several approaches
\citep{mikolov2013linguistic,luong2015bilingual,gouws2015bilbowa,sogaard2015inverted,xing2015normalized,artetxe-labaka-agirre:2016:EMNLP2016,mogadala2016bilingual,vulic2016bilingual,mikolov2013exploiting,zhang2016ten,faruqui-dyer:2014:EACL}. Many
methods require bilingual corpora aligned at the document, paragraph or
sentence level \citep{gouws2015bilbowa,luong2015bilingual,
  sogaard2015inverted,vulic2016bilingual,mogadala2016bilingual}, which can be difficult to obtain, particularly when dealing with low resource languages. In this work we focus on methods that rely on monolingual corpora and relatively small bilingual
dictionaries (containing ca. 5000 pairs) \citep{mikolov2013linguistic,artetxe-labaka-agirre:2016:EMNLP2016,xing2015normalized}, which are more widely available. A common strategy of these methods is to exploit structural similarities among separate vector spaces. For instance \citep{mikolov2013linguistic,mikolov2013exploiting}, 
started from two monolingual spaces, and then learned a linear transformation from the source embedding space to the target embedding space that minimizes the eucledian distance of the word pairs in the bilingual dictionary.
%The first attempt in this direction was implemented by  by learning a linear transformation which
%minimized Euclidean distances for word pairs in dictionaries.
This approach was improved by \citep{zhang2016ten}, who constrained the
transformation matrix to be orthogonal. Further, \citep{xing2015normalized} added length normalization and the maximization of the cosine similarity,  forcing the orthogonality constraint to keep the normalization after the mapping. While the latter authors learned
a single transformation from target to source language, \citep{faruqui-dyer:2014:EACL} used canonical correlation analysis to learn separate
transformations, projecting the monolingual spaces into a shared space.
More recently,  \citep{artetxe-labaka-agirre:2016:EMNLP2016} presented a framework that generalized previous approaches and showed
the common aspects among them.
%Moreover, \citep{artetxelearning} showed some theoretical flaws in their justifications, proposing alternative
%interpretations to them.
Their method includes orthogonality constraints, a global preprocessing with length normalization and dimension wise
mean centering, achieving the best accuracy to date in the dictionary induction task. We will use this latter method as a baseline. Note that these methods learn monolingual embeddings first, and then map those embeddings. Our method learns a joint bilingual space in one go. %Section  \ref{sec:exp} reports results of \citep{artetxelearning} head to head ours.   %% elebakarreko degradazioa aipatu!?

Finally, NASARI \cite{camacho2016nasari} combines text-based and knowledge-based methods, yielding multilingual vector representations. They exploit BabelNet  \cite{navigli2012}, a large multilingual semantic network that includes, among others, information from wordnets, Wikipedia, OmegaWiki and Wiktionary. They learn representations for BabelNet concepts with a method that analyzes the content words in Wikipedia articles, weighting them using a schema similar to \emph{tf-idf}, and mapping each word to its candidate senses according to BabelNet. As a result, each BabelNet concept is represented by a vector indexed by language-independent concepts, with excellent results in cross lingual similarity experiments. Although NASARI uses a richer knowledge base (BabelNet), we will compare the performance of NASARI head-to-head our method, which only uses wordnets.

\section{Bilingual embeddings via constraints}
\label{sec:constraints}

In this section we describe our method to learn bilingual embeddings using constraints. We start by reviewing the general skipgram model and then describe the extension to infuse information from bilingual dictionaries as regularizers in the loss function .

% In order to learn bilingual embeddings based on monolingual corpora
% and a bilingual dictionary, previous work has learned separate monolingual embeddings and a separate mapping function. We propose instead  to learn bilingual embeddings jointly. Our method merges the monolingual corpora, shuffling them at random, and infuses information from bilingual dictionaries, encoding them as bilingual constraints in the loss function of skipgram \citep{mikolov2013efficient}, one of the most effective methods to learn embeddings. We first review skipgram and then the extension to include bilingual constraints as regularizers.

\subsection{Skipgram}
\label{sec:loss}

The skipgram model \citep{mikolov2013efficient} uses the current word $w$ to predict the surrounding window of context words $c$. The parameters $\theta$ that maximize $p(c|w;\theta)$ are learned from a text corpus\footnote{The actual formula has been shown to be slightly different, but this is not relevant for the present work. We refer the interested reader to \cite{levy2014neural} for more details}. We will focus on the negative sampling implementation, where the loss function rewards the estimate of the probability for words that co-occur with each other, while penalizing the estimate of the probability for random word pairs co-occurring with each other. Negative sampling has shown to be a very effective alternative to the computationally expensive softmax, achieving state of the art results in many NLP tasks and substantially speeding up the learning
process.

Skipgram assigns two vectors\footnote{In the context of this article, we use word vectors and word embeddings interchangeably.} to each word in the vocabulary, which represent its semantic features as a target word and as a context word. In
this article we will refer to the word and context vector spaces as $W$ and $C$ respectively, which are the parameters $\theta$ to be learned. Both $W$ and $C$ are $|V| \times D$ size matrices, being $|V|$ the size of the vocabulary and $D$ the dimensionality of the vectors. Equation \ref{eq:losslng} describes
the loss function for each word-context occurrence in the corpus according to the skipgram model with negative sampling, being $w\in W$ the target word, $c\in C$ a context word and $c_n\in C$ a random negative sample (indexed by $n$) drawn from a noise distribution $P(c)$. 

\begin{equation}
\label{eq:losslng}
    %J_{sg}(w) = \sum_{k=1}^{K} ( log(\sigma(c_k^t w)) + \sum_{n=1}^{N} log(\sigma(-w^t_k c_n)))
    J_{sg}(w,c) = log(\sigma(c^t w)) + \sum_{n=1}^{N} \mathbb{E}_{c_n \sim P(c)} \bigg[log(\sigma(-c_n^t w))\bigg]
\end{equation}

%The leftmost term in the loss function models the probability of $w_k$ and $c$ to occur together, while the rightmost term models the probability of $w_k$ and a randomly chosen word. imize the loss function, balancing between rewarding the dot-product of the vectors of co-occurring word pairs ($w_k^t c$ in the leftmost term) and penalizing words that do not co-occur together ($-w^t_k c_n$ in the rightmost term). 

The total loss is the summation of Eq. (\ref{eq:losslng}) for all pairs of words $w$ and  $c$ co-occurring in the corpus, as extracted using a context window of size $K$. The model has, among others, hyperparameters  $K$ (context window size) and $N$ (number of negative samples). Stochastic gradient descent is used to find the parameters $W$ and $C$ that maximize the loss function in the corpus. 

%, so that the resulting gradients for a given $c$ and
%$w_k$ pair are as follow:
%\begin{equation} \label{eq:derC}
%\frac{\partial J_{sg}}{\partial c} = (1-\sigma(w_k^t c)) w_k
%\end{equation}
%\begin{equation} \label{eq:derWk}
%\frac{\partial J_{sg}}{\partial w_k} = (1-\sigma(w_k^t c)) c - \sum_{n=1}^{N} (1-\sigma(-w^t_k c_n)) c_n 
%\end{equation}
%\begin{equation}\label{eq:derCn}
% \frac{\partial J_{sg}}{\partial c_n} = -(1-\sigma(-w^t_k c_n)) w_k
%\end{equation}
%Equation \ref{eq:derC}, \ref{eq:derWk} and \ref{eq:derCn} are going to be used for updating $c$, $w_k$ and all $c_n$, respectively, suming up
%$K+K \times N+1$ derivatives per gradient descent.

The embeddings learned by skipgram have proven to be very effective modeling semantic and syntactic phenomena. For instance, the word vectors  in space $W$ have been used to model word equivalence and similarity, using the cosine of the vectors of two words to capture how similar the words are \cite{baroni2014don}. Evaluation of the word embeddings in $W$ in standard tests against human-curated gold standards has confirmed their quality \citep{baroni2014don}. In addition to using the embeddings in $W$ alone, other work has shown that both $W$ and $C$ contain complementary information \citep{pennington2014glove,levy2015improving}. Following their work, we represent a word as the addition of the respective two vectors in  $W$ and $C$, as preliminary experiments confirmed that these representations were more effective than using $W$ or $C$ alone. 

\subsection{Introducing bilingual constraints}
\label{sec:biconst}

%\todo[color=green!40]{[45] artikuluko lexical borrowing aipatu!?}

%\todo[color=green!40]{[45] artikuluko lexical borrowing aipatu!?} We propose a joint method that merges and shuffles two monolingual corpora for two languages, using skipgram on the merged corpus to learn similar embeddings for translation equivalents. This method depends on how strings are shared across languages (e.g numbers, proper names or loan words), as these shared strings allow to gather co-occurrences of mixed pairs, where one string in the pair belongs to both languages and the other to just one (e.g. \emph{Paris} -- \emph{beautiful} in English and \emph{Paris} -- \emph{polita} in Basque). Given enough such shared strings, skipgram will assign similar embeddings for \emph{beautiful} and \emph{polita}. Unfortunately, the number of such shared strings is usually too scarce and translation equivalents will tend to be distant in the embedding space, as we show empirically in the experimental section.

Given (unaligned) monolingual corpora in two languages, a straightforward method to obtain bilingual embeddings is to merge and shuffle the corpora at a sentence level, and then to use skipgram to learn embeddings on the merged corpus. The efficiency of this method critically depends on how strings are shared across languages (e.g numbers, proper names or loan words), as these shared strings act as bridges that allow gathering bilingual co-occurrences \cite{wick2016minimally}. Unfortunately, the number of such shared strings is in general too small, and translation equivalents tend to be distant in the embedding space, as we show empirically in the experimental section.

One way to reinforce translations in the joint embedding space is to include a regularization term in the loss function (Eq. \ref{eq:losslng}) that forces embeddings of translation equivalences to eventually become closer in the joint vector space. This regularization term was first introduced in \citep{halawi2012large} and \citep{bollegala2016joint} to infuse information derived from external sources into the learning process (synonyms, related words, etc.), and we apply it here to deal with translations gathered from a bilingual dictionary.

Equation \ref{eq:losslng2} shows the new loss function for each word occurrence $w$, which contains two terms: the loss function for co-occurrence pairs in Equation (\ref{eq:losslng}), and the regularizer which minimizes the L2 norm between $w$ and its translation equivalents $w_{lm}$ according to $\lambda \in \mathbb{R^+}$, the regularization coefficient. Translation equivalents for a word are indexed by $l$ (language) and $m$ (which is one of multiple $M_l(w)$ equivalents in that language). We also use monolingual constraints, where $l$ is the same language as that of word $w$. In this case we obtain synonyms of $w$ instead of translations equivalents. Both monolingual and bilingual constraints are extracted from wordnets (cf. section \ref{sec:exp1}).

\begin{equation}
\label{eq:losslng2}
    J_{sg+}(w) = \sum_{c \in window_K(W)}J_{sg}(w,c) - \lambda \sum_{l=1}^{2} \sum_{m=1}^{M_l(w)}|| w - w_{lm} ||^2_2
\end{equation}

\section{Random walks over knowledge bases}
\label{sec:rw}

As mentioned in the related work section, one of the most effective methods to build word representations from knowledge bases is to perform random walks over the knowledge base graph to collect co-occurrence data which reflects the structure of the knowledge base \citep{goikoetxea2015random}. Given the co-occurrence information, skipgram is used to produce embeddings that encode the meaning of words according to the knowledge base. Those co-occurrences, and derived embeddings, can be seen as the knowledge-based counterpart of text co-occurrences, and they characterize the meaning of a word from the perspective of the relations in the knowledge base. We will first present this method, and then extend it to represent words in two languages in the same embedding space.

%One of the key resources of the research in this paper are the pseudo-corpora created by random walks over WordNet. We have used the method
%proposed by \citep{goikoetxea2015random} in which the authors created synthetic contexts from WordNet running a Monte Carlo method for calculating the
%PageRank algorithm \citep{Avrachenkov:2007:MCM:1272804.1272825} with the publicly available UKB\footnote{http://ixa2.si.ehu.eus} tool.

  \begin{algorithm}[t]
   \caption{Monolingual Random Walks (MoRW)}
   \label{alg:morw}
    \begin{algorithmic}%[1]
      \State  \textbf{Input:} $C$ set of concepts\\
      				 \tab $N(c)$, the neighbors of concept $c \in C$ in the graph\\
      				 \tab $D(c)$, the lexicalizations of concept $c\in C$\\
%that maps the \textit{V} vertices to \textit{L} \\\tab languages and \textit{w} words\\
      				 \tab $I$, number of synthetic contexts\\
      				 \tab $\alpha$, damping factor
      \State \textbf{Output:} SC, monolingual synthetic corpus 
%      \Function{BiRW}{$G, L, D, S, \alpha$}
        %\State Let $M$ be a $N\times N$ transition probability matrix, where $M_{ij} = \frac{1}{d_{i}}$ if a link from $i$ to $j$ exists, and zero otherwise.
      \State $\mathrm{SC} \leftarrow []$
      \State $i\leftarrow 0$
       \Repeat
%       \For{\textit{1..I}} 	
            \State $S \leftarrow []$
            \State Choose vertex $c\in C$ with prob. ${1}/{|C|}$
            \Repeat \Comment{Keep walking}
              \State Choose word $w\in D(c)$ with prob. ${1}/{|D(c)|}$ and push it to S
              \State Choose vertex $c^{\prime} \in N(c)$ with prob. ${1}/{|N(c)|}$
              \State $c \leftarrow c^{\prime}$
            \Until {$\mathrm{random}() > \alpha$} \Comment{Halt walk}
            \State $\mathrm{SC} = \mathrm{SC} \cup S$ \Comment{New context}
            \State $i \leftarrow i + 1$
%            \While {$random(1) > (1 - \alpha)$} \Comment{Keep walking}
%            	\State Choose $c'\in neighbors(c)$ with probability ${1}/{|neighbors(c)|}$
%                \State Choose language $l\in L$ with probability $0.5$ %$\frac{1}{L}$
%                \State Choose word $w\in D_l(c')$ with probability ${1}/{|D_l(c')|}$ and emit it in PC
%                \State Choose word $w\in D_l(c')$ with probability ${1}/{|D_l(c')|}$ and emit it in PC
%            \EndWhile							
        \Until {$i == I$}\\
%        \EndFor\\
%        \Return    PC
%       \EndFunction
\end{algorithmic}
\end{algorithm}
%\end{enumerate}

\subsection{Monolingual random walks}

Let's model the target knowledge base as a graph $G = (C,E)$, being $C$ the list of vertices (concepts) and $E$ the edges between concepts (relations in the knowledge base). In addition,
let $N(c)$ be the set of neighbors of node $c$ in the graph, and let $D(c)$ be the possible lexicalizations of concept $c$.

% we have a dictionary listing possible meanings in $C$ for each word in the vocabulary $V$ of the knowledge base. For instance the word \textit{paw} is linked to concepts \textit{02439929-n 05564590-n 01211455-v 01211098-v} in wordnet. We also have the inverse dictionary, which lists, for each concept its lexicalizations. For instance the concept \textit{02439929-n}  is lexicalized with the English word \textit{paw} in the English WordNet and the Basque words \textit{erpe, hatzapar} and \textit{hatz} in the Basque WordNet. 

Random walks are simulated by means of a Monte Carlo method, as follows. The algorithm starts at one random concept from $C$. Every step of the walk the algorithm tosses
a coin: with probability $\alpha$ a neighbor concept is selected following one edge at random; with probability $1-\alpha$ the walk is halted and a new concept from $C$ is chosen at random. Every time a concept is selected, a possible lexicalization is chosen and printed. Every time the walk is halted, a new context is produced (that is, a newline is printed). The algorithm stops once it reaches a pre-fixed number of walks $I$. The algorithm has thus two hyperparameters, $\alpha$ and $I$. All probability distributions and random choices, except $\alpha$, follow the uniform distribution. Algorithm \ref{alg:morw} summarizes the procedure to perform monolingual random walks. 

For instance, a synthetic context created out of WordNet looks like the following: 

\begin{center}
\emph{paw feline felid ounce panthera lion sekhet}
\end{center}

%The latter context has been extracted from WordNet 3.0 with glosses\todo{nonbaiten azaldu behar dira wordnet eta wordneteko erlzioak}, and it starts with the $gomphothere$, an extinct elephant of Prehistoric
%Ages; then jumps to $glacial\_epoch$, which is an specific Prehistoric Age; afterwards emmits $sabertooth$, a big extincted feline of the Ice
%Age; the next word is $panthera$, a contemporary feline; and finally halts with the more abstract $mammal\_genus$, the term used in Biology for
%designating the whole group of mammals. 

The latter context has been produced using WordNet 3.0 with gloss relations and $\alpha=0.85$. The lexicalization for the first vertex is \emph{paw}, a soft foot of mammals which normally has claws; the walk continues following an edge to the more abstract \emph{feline}, a subfamily of cats that includes mostly small to medium-sized cats; follows another edge to  the also abstract \emph{felid}, the family of cats that the previous word \emph{feline} belongs to; follows a hyponymy edge to  \emph{ounce},  a long haired leopardlike feline of the mountains of central Asia, also known as snow leopard; then comes \emph{panthera}, the genus that \emph{ounce} belongs to; follows another edge to the concept lexicalized by \emph{lion}, also related to the genus \emph{panthera}; continues with the Egyptian lion headed goddess \emph{sekhet} via a gloss relation, and finally halts the walk. The algorithm would continue until the desired number of contexts $I$ is obtained.

%Ages; then jumps to $glacial\_epoch$, which is an specific Prehistoric Age; afterwards emmits $sabertooth$, a big extincted feline of the Ice
%Age; the next word is $panthera$, a contemporary feline; and finally halts with the more abstract $mammal\_genus$, the term used in Biology for
%designating the whole group of mammals. 

%This example gives an insight about the nature of the semantic information coded in the pseudo-corpora
%in an latent manner, due to the fact that neighbor concepts in a KB are lexicalized following the relations among the concepts and thus
%following the structure of the KB. 

% As mentioned in section \ref{sec:relwork}, \citep{goikoetxea2015random}  proved that the synthetic corpus
% encodes WordNet relations as co-occurrences, and that embeddings learned from this synthetic corpus provide excellent results in word similarity tasks.

This synthetic corpus alone is a good resource to learn embeddings from, as it leverages the structural information of WordNet in the form of word co-occurrences in the synthetics corpus, and provides excellent results in monoingual word similarity tasks \citep{goikoetxea2015random}. But, given the complementarity of the information in knowledge bases and text corpora, it is possible to combine knowledge-based and distributional information with excellent results  \citep{goikoetxea2016single}. We follow their method, which concatenates and shuffles the synthetic corpus with natural text corpora at the sentence level, and then learns embeddings from this hybrid corpus. For instance, the method would mix co-occurrences from the above example for \emph{panthera} with those of naturally occurring text as the following:

\begin{center}
\emph{Felids of the genus Panthera (tigers, lions, jaguars and leopards)\\ also produce sounds similar to purring.}
\end{center}

%\todo{bi paragrafo hauek berridazteko daude} If a NNLM processes that kind of hybrid corpus, when computing the embedding of a word that appears in both text and pseudo-corpus
%the algorithm takes into account the co-occurrences with neigbors from both sources. In other words, the embeddings of words that co-occur with
%neighbors from natural language text and from knowledge bases enclose semantic information from both. Lets take a look to the word \emph{panthera} in the next
%example: 

%While the words that co-occur with \emph{panthera} in the latter sentence respond to semantic as well as syntantic relations within a grammatical
%sentence, the neighbors from the synthetic context appear along with \emph{panthera} because of relations among concepts within the structure
%of WordNet. Thus the NNLM merges \emph{panthera}'s semantic information of neighbors coming from both sources into a single embedding, creating
%a more complete meaning.

%Note that \citep{goikoetxea2016single} author's proposal has been implemented just for English, and that its goal is to enrich a monolingual
%vector space. In \ref{sec:birw} we explain how we have extended the random walks from monolingual to bilingual, so that the resulting pseudo-copora
%gives way to construct a compact bilingual vector space.

\subsection{Bilingual random walks}
\label{sec:birw}

%In this paper we propose to extend knowledge-based random walks to bilingual wordnets, emitting both words at random to represent words in both languages in the same embeddings space. 
The method for generating a bilingual synthetic corpora is the same as for the monolingual case, but, when arriving to a concept, a lexicalization in any of the two languages is emitted at random. The method is formalized in Algorithm \ref{alg:birw}, where $C$ and $N(c)$ are defined as before, and $D_{l}(w)$ is the set of all lexicalizations of concept $c$ in language $l \in \{l_A, l_B\}$. Note that $D_{l}(c)$ can be empty when the concept is not lexicalized in that particular language, due to different coverages of the wordnets or lexicalization differences across languages. Algorithm \ref{alg:birw} thus controls for this case and also for potentially empty contexts. % Note that nothing is emitted when a concept is not lexicalized in the chosen language $l$ (that is, $D_l(c)$, returns the empty set).

%In section \ref{sec:rw} we have explained in detail how to extract pseudo-corpora out of a KB (e.g. WordNet), the nature of the semantic information in the pseudo-corpora and
%their embeddings, as well as the complementariness of data in text and KB pseudo-corpora. Given that the latter resources and methods have shown to be very successful in
%monolingual similarity tasks, we have extended it to more than a language in order to enhance the results in cross-lingual similarity.

%whether the wordnet  which are the possible lexicalizations. synsets and relations among them are the key for
%linking languages in that they constitute the common semantic net beyond lexicalization, and thus they can join not only words within a language but across languages. 
%In fact, lexicalization of some languages are mapped in the same semantic net (e.g. English and Basque), or, in the case of languages with different nets, lexicalizations are joined by the mapping across synsets (e.g. English and Italian). 

%\begin{enumerate}

  \begin{algorithm}[]
   \caption{Bilingual Random Walks (BiRW)}
   \label{alg:birw}
    \begin{algorithmic}%[1]
      \State  \textbf{Input:} $C$ set of concepts\\
      				 \tab $N(c)$, the neighbors of concept $c \in C$ in the graph\\
      				 \tab $D_l(c)$, the set of lexicalizations of $c\in C$ in language $l\in \{l_A,l_B\}$\\ 
%that maps the \textit{V} vertices to \textit{L} \\\tab languages and \textit{w} words\\
      				 \tab $I$, number of synthetic contexts\\
      				 \tab $\alpha$, damping factor
      \State \textbf{Output:} SC, bilingual synthetic corpus 
%      \Function{BiRW}{$G, L, D, S, \alpha$}
        %\State Let $M$ be a $N\times N$ transition probability matrix, where $M_{ij} = \frac{1}{d_{i}}$ if a link from $i$ to $j$ exists, and zero otherwise.
      \State $\mathrm{SC} \leftarrow []$
      \State $i\leftarrow 0$
       \Repeat
            \State $S \leftarrow []$
            \State Choose vertex $c\in C$ with prob. ${1}/{|C|}$
            \Repeat \Comment{Keep walking}
              \State Choose language $l\in \{l_A,l_B\}$ with prob. $\frac{1}{2}$ %$\frac{1}{L}$
              \If {$|D_l(c)| > 0$}
              \State Choose word $w\in D_l(c)$ with prob. ${1}/{|D_l(c)|}$ and push it to $S$ \EndIf
              \State Choose vertex $c^{\prime} \in N(c)$ with prob. ${1}/{|N(c)|}$
              \State $c \leftarrow c^{\prime}$
            \Until {$\mathrm{random}() > \alpha$} \Comment{Halt walk}
            \If {$|S| > 0$}
              \State $\mathrm{SC} = \mathrm{SC} \cup S$ \Comment{New context}
              \State $i \leftarrow i + 1$
            \EndIf 
%            \While {$random(1) > (1 - \alpha)$} \Comment{Keep walking}
%            	\State Choose $c'\in neighbors(c)$ with probability ${1}/{|neighbors(c)|}$
%                \State Choose language $l\in L$ with probability $0.5$ %$\frac{1}{L}$
%                \State Choose word $w\in D_l(c')$ with probability ${1}/{|D_l(c')|}$ and emit it in PC
%                \State Choose word $w\in D_l(c')$ with probability ${1}/{|D_l(c')|}$ and emit it in PC
%            \EndWhile							
        \Until {$i == I$}\\
%        \Return    PC
%       \EndFunction
\end{algorithmic}
\end{algorithm}
%\end{enumerate}

There are currently open wordnets in thirty-four languages \footnote{http://compling.hss.ntu.edu.sg/omw/}. Some of them have been created following the original
English structure (\emph{expand} approach), where the concepts are lexicalized in more than one language and the relations between concepts are kept untouched. Other wordnets have been created separately (\emph{merge} approach), with different sets of concepts and relations, and a mapping to the English WordNet is provided. In the experiments (cf. Section \ref{sec:exp}) we will use the English WordNet, two expand-approach wordnets (Basque and Spanish) and one merge-approach wordnet (Italian). Note that the wordnets vary in size, with English having the largest and Basque the smallest (see Table \ref{tab:wn}). 

Wordnets created using the expand approach share the concept inventory and the relations between those concepts, and thus cross-lingual information is fully compatible. Ideally, all concepts should be lexicalized in both languages, but in reality only a fraction of the concepts is lexicalized in both languages, depending on the sizes the  wordnets involved. %In the case of merge wordnets, the coverage of the mapping can also affect how many lexicalizations are accesible. 
In the merge approach, a mapping between wordnets is necessary to find equivalent concepts and relations across languages. In the case of the Italian WordNet, we added all concepts of the Italian and English wordnets to the graph, as well as the monolingual relations in each wordnet, plus the mapping between Italian and English concepts.
%\todo{italieraren kasuan komentatu bi grafo daudela, mapaketa erlazio bezala sartuta. Kasu horretan synset batzuen hiztegian ingelesa edo italiera dago. Hori dela eta algoritmoak portaera ezberdina edukiko du (\%50ean ez du ezer emetituko): hizkuntza bateko asko jarraian, zubia pasatzean italianora, baina tartean emititu gabeko synsetak.}

%, as in section \ref{sec:rw}, but with the following differences: it uses a bilingual dictionary (see \ref{sec:exp1}) mapped in the same graph, and if the languages have separated WordNets a double graph.

%As well as in the monolingual random walks (see section \ref{sec:rw}), in the bilingual ones the walks' halts and jumps are determined by $\alpha$ in the very same way. But when the walk reaches a node, the inverse dictionary links the node to words of more than a language and thus it randomly emits any of them. It goes without saying that the more lexicalizations the language has the more probabilities its word to be emitted. 

The next example shows a bilingual synthetic context created for English and Basque (in bold) which includes \emph{panthera}:

\begin{center}
 \emph{\textbf{elur-pantera} panthera felidae sabertooth smiledon \textbf{ugaztun} erignathus}
\end{center}

The example starts with the Basque word \textit{elur-pantera} (\textit{snow leopard}); then jumps to the more general lexicalization in English \textit{panthera}, remains in the same language with the even more general \textit{felidae} and goes more specific again in \textit{sabertooth} and \textit{smiledon}; in the next step it emits the abstract Basque word \textit{ugaztun} (\textit{mammal} in English); the walk halts in \textit{erignathus}, a term that refers to the family of seals. The unbalance of lexicalizations in Basque and English is reflected in the bilingual random walks, with more lexicalizations available for English. %, and, evidently, it remains the same all over the pseudo-corpus.  As we will see in \ref{sec:exp}, in this research, the token unbalance among languages is a fact we had to take into account in order to control the number of tokens coming from each language when creating balanced bilingual copora.

Similar to monolingual random walks, the bilingual random walks can be combined with monolingual text corpora, producing a bilingual hybrid corpora which mixes bilingual co-occurrences from bilingual random walks and monolingual co-occurrences from the monolingual corpora. We will see that this is an effective method to produce bilingual embeddings. 

%Note that in bilingual pseudo-corpora words from different languages co-occur with each other following semantic relations in WordNet. Thus besides containing the relations in knowledge bases implicitly as commented in \ref{sec:rw}, bilingual pseudo-corpora relate words across languages. The latter implies that if a bilingual pseudo-corpora is processed by an NNLM, words that co-occur across languages will have similar representations. This unification of the bilingual data within a vector space is due to the latent structure of the pseudo-corpora, being the main contribution of the bilingual random walks. So, the potential of bilingual random walks relies upon the semantic relations between concepts, which trascend word forms and allow words across languages to co-occur with each other.

%Further, in section \ref{sec:exp} we will prove that, on the one hand, the complementariness of semantic information in text and knowledge bases described in \ref{sec:rw} also works in a bilingual space; on the other, that bilingual random walks are complementary to bilingual constraints. 
%in that the combination of both reinforces the similarity across languages.

\section{Experiments}
\label{sec:exp}

Word similarity and relatedness are the most common evaluation methods to measure the quality of monolingual embeddings \citep{agirre2009study,luong2013better,baroni2014don,levy2015improving,faruqui-EtAl:2015:NAACL-HLT}, and we thus chose cross-lingual word similarity and relatedness to measure the quality of bilingual word embeddings. In cross-lingual word similarity and relatedness, given a set of pair of words in two languages, the goal is to return scores which are close to human-curated gold standard ground truth scores. %similarity and relatedness scores.
Given two words, the automatic score is obtained by computing the cosine similarity of the corresponding embedding vectors.  The evaluation measure is typically Spearman's rank correlation\footnote{\url{https://en.wikipedia.org/wiki/Spearman\%27s_rank_correlation_coefficient}}. 

% It is difficult to annotate cross-lingual gold standard datasets directly, as volunteers that are proficient in both languages are not easy to recruit. Instead, \citep{camacho2015framework} proposed a method to construct a cross-lingual dataset based on a monolingual dataset which is available in more than one language. The method consists, basically, in using the average of the monolingual scores. For instance, \citep{camacho2015framework} have translated RG, the popular English similarity dataset \citep{rubenstein1965contextual}, into Spanish, also collecting scores for the Spanish pairs. Given an English pair like \emph{journey} -- \emph{voyage} with score $3.58$ (comprised within $[0..4]$), it was translated as \emph{trayecto} -- \emph{viaje} and scored with $3.13$. As common with out of context translations, \emph{recorrido} is not an exact equivalent of \emph{journey}, which is reflected in the different scores. Given these two pairs, two cross-lingual pairs are produced, with mean score $3.36$: \emph{journey} -- \emph{viaje} and \emph{voyage} -- \emph{trayecto}. Please refer to \citep{camacho2015framework} for a detailed description of this method.

We  reviewed  existing word similarity and relatedness datasets in a number of languages, with English having the highest amount of datasets. In order to select the target languages we took into account the following factors: the existence of at least two datasets for the language (see next Section) and the existence of a sizeable open wordnet\footnote{\url{http://compling.hss.ntu.edu.sg/omw/}}. The overlap yielded three languages, English, Italian and Spanish, two romance and one west germanic language. In order to add more linguistic diversity, we selected Basque, an agglutinative language which is a language isolate (no related language is known). There is a sizeable wordnet for Basque, and we manually annotated two datasets. As we will see in detail below, our experiments involve four languages and twelve cross-lingual datasets in six language pairs: English-Basque (ENEU), English-Spanish (ENES), English-Italian (ENIT), Spanish-Basque (ESEU), Spanish-Italian (ESIT) and Italian-Basque (ITEU). 

\subsection{Experimental setting}
\label{sec:exp1}

In order to learn the embeddings for the target languages we have used readily available \textbf{Wikipedia  corpora}\footnote{\url{http://linguatools.org/tools/corpora/wikipedia-monolingual-corpora/}}, except for Basque, which was not available. The Basque Wikipedia corpus was extracted from the 07/04/2016 dump, and, given its smaller size, it was complemented with the Elhuyar Web Corpus \citep{leturia2012evaluating}, kindly made available from the authors on request. As Basque is an agglutinative language, we have stemmed the Basque corpus, but the rest of the corpora were not. All corpora were converted to lowercase. The sizes of the monolingual corpora are shown in Table \ref{tab:monotxt}. %Note that the Basque corpus is the smallest, less than half the size of the Italian corpus, and ten times smaller than the English corpus.

\begin{table}[t] %!htb]
\centering
\begin{tabular}{l|r|}
\rowcolor[HTML]{FFFFFF} 
\cline{2-2}
\multicolumn{1}{c|}{\cellcolor[HTML]{FFFFFF}} & \multicolumn{1}{|c|}{Tokens} \\ \hline
\multicolumn{1}{|l|}{EU}                      & $160 \cdot 10^6$     \\
\multicolumn{1}{|l|}{IT}                      & $380 \cdot 10^6$    \\
\multicolumn{1}{|l|}{ES}                      & $430 \cdot 10^6$    \\ %\hline
\multicolumn{1}{|l|}{EN}                      & $1,700\cdot 10^6$   \\ \hline
\end{tabular}
\caption{Amount of tokens per monolingual corpora, in ascending order. }
\label{tab:monotxt}
\end{table}

\begin{table}[t]
\centering
\begin{tabular}{c|r|r|}
\cline{2-3}
\multicolumn{1}{l|}{}    & \multicolumn{1}{c|}{Words}  & \multicolumn{1}{c|}{Synsets} \\ \hline
\multicolumn{1}{|c|}{EU} & 26,701  & 30,464   \\ %\hline
\multicolumn{1}{|c|}{IT} & 46,679  & 49,515   \\ %\hline
\multicolumn{1}{|c|}{ES} & 53,039  & 55,814   \\ %hline
\multicolumn{1}{|c|}{EN} & 147,306 & 136,334  \\ \hline
\end{tabular}
\caption{Number of words and synsets in each wordnet, in ascending order.}
\label{tab:wn}
\end{table}

\begin{table}[t]
\centering
\begin{tabular}{|l|c|c|c|l|c|l|}
\hline
%\cline{2-7}
\multicolumn{4}{|c|}{\cellcolor[HTML]{FFFFFF}monolingual} & \multicolumn{2}{c|}{bilingual} & \multicolumn{1}{c|}{total} \\ \hline
EU         & 24,985          & IT         & 33,608        & EUIT         & 109,031         & 167,624                    \\
EU         & 24,985          & ES         & 43,218        & EUES         & 67,685          & 135,888                    \\
ES         & 43,218          & IT         & 33,608        & ESIT         & 126,149         & 202,975                    \\
EN         & 152,219         & EU         & 24,985        & ENEU         & 99,861          & 277,065                    \\
EN         & 152,219         & IT         & 33,608        & ENIT         & 173,126         & 358,953                    \\
EN         & 152,219         & ES         & 43,218        & ENES         & 176,873         & 372,310                    \\ \hline
\end{tabular}
\caption{Amount of pairwise constraints per language pair, in ascending order. For each language pair, we report number of monolingual and bilingual constraints alongside the total.}
\label{tab:constnum}
\end{table}

Regarding \textbf{wordnets}, we have used the Basque, English and Spanish open wordnets corresponding to the Multilingual Central Repository 3.0 \citep{agirre2012multilingual}. Both the Basque and Spanish wordnets reuse the synsets\footnote{In WordNet, concepts are referred to as synsets. We use both terms interchangeably in this article.}, hierarchy and relations from the English WordNet 3.0 version, including the gloss relations \footnote{http://wordnet.princeton.edu/glosstag.shtml}. For Italian we used ItalWordNet, which has its own set of synsets and relations \citep{iwn03}, and includes links to the English WordNet 3.0. All are available from the Open Multilingual WordNet website\footnote{\url{http://compling.hss.ntu.edu.sg/omw/}}. Table \ref{tab:wn} reports the number of words and synsets in each wordnet. All lexicalizations have been lowercased for compatibility with the corpora.

%Our random walk algorithm assumes that the synset inventory is common for both languages,\todo{aurrerago, lotu ondo 3 ataleko azalpenekin} which is not true for Italian. In the case of Italian, we used the available mapping to move the lexicalization information to the English synsets, that is, we took the English synset inventory and added Italian lexicalizations based on the mapping

%It is important to emphasize the relevance of the inverse dictionaries in the above pairs of languages, since they are the visible face of the underlying framework
%upon which their respective bilingual constraints are built on and the bilingual random walks depend on. For pairs which share the same synset structure (ENEU and ENES) the reference for bilingual constraints and random walks are going to be concatenated dictionaries. In the case ENIT (separate synset stuctures) we had to map the Italian WordNet's synsets onto the English ones before concatenating them. That way, each pair will have its lexicalizations in two languages linked to a common synset structure in their respective bilingual dictionaries. 

In order to produce the \textbf{bilingual constraints}, %in the same manner as the monolingual synonyms in \citep{halawi2012large,bollegala2015joint}, 
we mined the dictionaries in the wordnets. That is, for each entry in the vocabulary, we checked its synsets and produced all possible lexicalizations in the same language (monolingual synonyms) and bilingual translation equivalents. Table \ref{tab:constnum} reports the number of pairwise constraints for each language pair.

%\todo{ENES constraint kopurua gaizki dagoela esango nuke} Algoritmoaren inguruan \todo[color=green!40]{Ondo dago... Elebakar guztiak ingelesezkoak dira}

%Given that \citep{halawi2012large} used only synonymy and \citep{bollegala2015joint} author's overall results were attained with the same relation, our bilingual constraints
%have been created upon synonymy across languages. Thus, for every entry in the bilingual dictionary we assigned all synonymy-related words in both languages; that is, for every
%entry, we checked its synsets and assigned all their lexicalizations in both languages. In relation to the bilingual random walks (see section \ref{sec:birw}), the walks are going
%to be performed over a bilingual inverse dictionary, all of its lexicalizations being mapped in the English synset structure as mentioned in this paragraph.

%Note the unbalance in the size of text corpora and WordNet that present Basque, Spanish and Italian w.r.t. the English, since it is going to be an important factor for controlling the contribution of each language when creating bilingual corpora in the following sections.

Regarding the evaluation datasets, the available \textbf{cross-lingual datasets} were the cross-lingual counterparts of the following well-known monolingual  datasets: a word relatedness dataset, WordSim353 (WS) \citep{gabrilovich2007computing}, and two similarity datasets, SimLex999 (SL) \citep{hill2016simlex} and RG \citep{rubenstein1965contextual}.
%Note that while the latter two have been constructed as strict  similarity datasets, WordSim353 mixed similarity and relatedness when gathering human scores.
In order to build the cross-lingual datasets, we followed the procedure proposed by \citep{camacho2015framework}, which combines two monolingual versions of the same dataset to produce a cross-lingual dataset. 
The monolingual versions that we used to construct the cross-lingual datasets are the following: the Italian versions of WordSim353 and SimLex999, as well as the RG version in Spanish created by \citep{camacho2015framework}; the Spanish WordSim353 and SimLex999 presented by \citep{hassan2009cross} and \citep{ETCHEVERRY16.1212}, respectively; the Basque WordSim353 and RG datasets produced in-house.  We thus produced  two datasets for each of the six language pairs, with the exception of English-Spanish, where we could produce three datasets, and Italian-Basque, with one common dataset, totaling twelve datasets.

%\begin{itemize}
%\color{red}
%\item TODO: EUSKARAZKO DATASETENTZAT atal bereizia????
%\end{itemize}

We did not fit skipgram hyperparameters, and reused the parameters reported by \citep{goikoetxea2016single} for calculating embeddings and the same regularization coefficient $\lambda$  as in \citep{halawi2012large}. In summary, dimensionality of 300, window size of 5, number of negative samples of 5, sub-sampling threshold of zero, and regularization coefficient $\lambda$ of 0.01. The damping factors for random walks was set to $\alpha=0.85$ following \cite{goikoetxea2015random}.

%\subsection{Results using bilingual constraints}
\subsection{Results using bilingual constraints on text corpora}
\label{sec:exp2}

In the first batch of experiments we focus on genuine text corpora, and check whether our method to learn embeddings jointly from monolingual corpora with bilingual constraints from wordnets is effective (\textsc{jointc}$_\text{txt}$ for short, cf. Section \ref{sec:constraints}). 
We shuffled  the sentences in the concatenation of two monolingual text corpora, yielding one bilingual corpus  for each of the  six language pairs, and then learned skipgram embeddings with the bilingual constraints extracted from the wordnets.
Due to the unbalanced sizes of the corpora accross languages (cf. Table \ref{tab:monotxt}), we experimented with two different combinations: equal number of tokens in both languages (reducing the size of the largest corpus in the pair with random selection), or all tokens in both languages. 
Preliminary results showed that the first option, with balanced corpus sizes, produced comparable or better results\footnote{In addition, we also tried with larger English corpora totaling $5\cdot 10^9$ tokens, with no improvement.}, so all experiments of the joint method in this article have been performed with balanced corpora.  
%Note that the balanced corpora are not aligned; moreover, the English subcorpora are created at random. Thus every corpus in this section and in further experiments are balanced w.r.t. the roken contribution of languages that comprise the pairs. 
Table \ref{tab:bitxt} shows the sizes of the balanced merged text corpora. Note that the bilingual text in the merged corpora is not aligned at all, and that each sentence comes from one language. As a baseline, we also learned bilingual embeddings with no use of constraints, \textsc{joint}$_\text{txt}$ for short.

We compare the results of our bilingual embeddings with  
a state-of-the-art method  \citep{artetxe-labaka-agirre:2016:EMNLP2016} (\textsc{map}$_\text{txt}$) which learns separate monolingual embeddings for each language using the corpora in Table \ref{tab:monotxt}\footnote{Note that this method has access to more tokens than our method, as better results were observed with larger corpora.},
and then learns a mapping based on the bilingual dictionary extracted from the wordnets  (cf. Section \ref{sec:relwork}). We use the same bilingual dictionaries as in our method (cf. Table \ref{tab:constnum}). 
The mapping is directional, we learned a mapping from the language with the smaller wordnet  to the language with the larger wordnet, obtaining six mappings, one for each language pair. In order to compute the similarity between two words, as customary \citep{artetxe-labaka-agirre:2016:EMNLP2016}, we first apply the learned mapping to one of the words.
%compute the similarity between the embeddings of\textbf{} the target language and the embeddings of source language after the mapping:
For instance, for a pair with an English word and a Basque word, we compute the cosine between the mapped embedding of the Basque word and the embedding of the English word. %\todo{Mikelekin konprobatu, uste dut berak kontrako norabidean egiten duela}
%In addition we also tried a basic baseline where the embeddings are compared with no mapping (MOTXT).

\begin{table}[t]%!htb]
\centering
\begin{tabular}{|l|c|}
\cline{2-2}
\multicolumn{1}{c|}{} & {Tokens} \\ \cline{1-2}
{ENEU}                      & $320 \cdot 10^6$    \\
{ESEU}                      & $320 \cdot 10^6$    \\
{EUIT}                      & $320 \cdot 10^6$    \\ 
{ENIT}                      & $760 \cdot 10^6$    \\
{ESIT}                      & $760 \cdot 10^6$    \\
{ENES}                      & $860 \cdot 10^6$    \\ \hline
\end{tabular}
\caption{Amount of tokens per merged text corpus, in ascending order. Each bilingual corpus comprises 50\% of monolingual sentences from each language.}
\label{tab:bitxt}
\end{table}

%Table \ref{tab:bic} reports the similarity results yielded when just combining text corpora in two languages (BITXT) and applying bilingual constraints (BIC) for the three language pairs, being compared to the results of the state-of-the-art method developed by \citep{artetxelearning} (MAP) (see \ref{sec:relw}) over separate monolingual vector spaces (MOTXT). %The latter method maps a source monolingual vector space into a target space optimizing a linear transformation among languages, using a bilingual dictionary as reference. 

%MAP is fed with separately learned monolingual vector spaces, and our method computes a single bilingual vector space out of a bilingual corpus. Thereby, in order to make both methods comparable, we have computed the separated vector spaces for MAP feeding word2vec with the same corpora with which we have created the bilingual corpora for out method. Further, the bilingual dictionary for MAP has been created out of the same bilingual constraints in our method.

\begin{table}[t]
\centering
\begin{adjustbox}{width=1\textwidth}
\begin{tabular}{|l|l|rrr|rrr|r|r|}
\cline{3-8}
\multicolumn{1}{c}{}      & \multicolumn{1}{c}{} & \multicolumn{1}{|c}{WS} & \multicolumn{1}{c}{SL} & \multicolumn{1}{c}{RG} & \multicolumn{1}{|c}{WS} & \multicolumn{1}{c}{SL} & \multicolumn{1}{c|}{RG} & \multicolumn{1}{l}{} & \multicolumn{1}{l}{}\\ \hline
%{}                       & MOTXT     & -4.1          & ---           & 3.9           & 10.5          & 3.8           & --- & {MOTXT} & {} \\                                                                                      
%{}                       & MOTXT+MAP & {67.2}                 & ---           & {78.8}                 & {60.5}        & {35.6}                 & --- & {MOTXT+MAP}  & {}   \\ %\cline{2-9}
{}                       & \textsc{map}$_\text{txt}$ 		& \textbf{68.5} & ---           & \textbf{81.5} & 59.8          & \textbf{35.8} & --- & {\textsc{map}$_\text{txt}$} & {}   \\ %\cline{2-9}                                       
{}                       & \textsc{joint}$_\text{txt}$     	& 52.8          & ---           & 58.3          & 57.6          & 24.9          & --- & {\textsc{joint}$_\text{txt}$} & {} \\                                                                                      
{\multirow{-3}{*}{ENEU}} & \textsc{jointc}$_\text{txt}$		& 61.6          & ---           & 70.9          & \textbf{62.5} & 26.9          & --- & {\textsc{jointc}$_\text{txt}$}  & {\multirow{-3}{*}{ENIT}} \\ \hline %\hhline{|==========|}%\hline                      
%{}                      & MOTXT     & 7.6           & 1.0           & 7.1           & -3.2          & -2.6          & --- & {MOTXT} & {}\\                                                                                       
%{}                      & MOTXT+MAP &        {63.0} &         {33.6} &        {61.3} &        {56.2} &        {31.9} & --- & {MOTXT+MAP}  & {} \\ %\cline{2-9}
{}                       & \textsc{map}$_\text{txt}$       & \textbf{62.6} & \textbf{33.9} & \textbf{79.5} & \textbf{55.9} & \textbf{31.6} & --- & {\textsc{map}$_\text{txt}$} & {} \\ %\cline{2-9}                                                                    
{}                       & \textsc{joint}$_\text{txt}$     & 59.0          & 20.5          & 63.8          & 50.6          & 25.1          & --- & {\textsc{joint}$_\text{txt}$} & {} \\                                                                                      
{\multirow{-3}{*}{ENES}} & \textsc{jointc}$_\text{txt}$    & 62.5          & 26.9          & 71.2          & 52.4          & 28.7          & --- & {\textsc{jointc}$_\text{txt}$}  & {\multirow{-3}{*}{ESIT}}                       \\ \hline %\hhline{|==========|}%\hline
%{}                      & MOTXT     & 6.6           & ---           & -1.4           & 3.8           & ---           & --- & {MOTXT} & {}\\                                                                                      
%{}                      & MOTXT+MAP &        {63.1} & ---           &        {61.9} &        {58.4} & ---           & --- & {MOTXT+MAP}  & {}\\ %\cline{2-9}
{}                       & \textsc{map}$_\text{txt}$ 		& \textbf{64.4} & ---           & \textbf{65.1} & \textbf{60.9} & ---           & --- & {\textsc{map}$_\text{txt}$} & {}   \\ %\cline{2-9}                                                                  
{}                       & \textsc{joint}$_\text{txt}$     	& 51.5          & ---           & 41.7          & 46.3          & ---           & --- & {\textsc{joint}$_\text{txt}$} & {}\\                                                                                       
{\multirow{-3}{*}{ESEU}} & \textsc{jointc}$_\text{txt}$ 	& {57.5}        & ---           & {49.0}        & {51.5}        & ---           & --- & {\textsc{jointc}$_\text{txt}$}  & {\multirow{-3}{*}{EUIT}} \\ \hline 
\end{tabular}
\end{adjustbox}
\caption{Results (Spearman) on the cross-lingual datasets using text corpora alone. \textsc{map}$_\text{txt}$ refers to monolingual text vector spaces with mapping. \textsc{joint}$_\text{txt}$ and \textsc{jointc}$_\text{txt}$ refer to joint bilingual vector space, without and with bilingual constraints, respectively. Best result for each dataset and language pair in bold.}
\label{tab:bic}
\end{table}

Table \ref{tab:bic} reports the results for the six language pairs.
%We can see that computing the similarity between independently learned monolingual embeddings (MOTXT) is close to zero in all cases, as expected.
The results of the baseline (\textsc{joint}$_\text{txt}$) are good, even if the text in one language is independent from each other. 
This results agree with those reported by \citep{wick2016minimally}, who attributed the surprisingly good results to \textit{lexical borrowing}, including numbers, loan words and named-entities, such as \emph{Hamlet} or \emph{Mandela}, that are shared across the two languages. It thus seems that the mere concatenation of texts already provides enough co-occurrence information to yield acceptable results when learning a joint embedding space. 

The method to infuse bilingual constraints  (\textsc{jointc}$_\text{txt}$) improves the results across all language pairs, showing that our method to modify the loss function of skipgram (cf. Section \ref{sec:biconst}) is effective. 
In any case, the mapping method from \citep{artetxe-labaka-agirre:2016:EMNLP2016} (\textsc{map}$_\text{txt}$) yields the best results across the board, except for ENIT, where \textsc{jointc}$_\text{txt}$ obtains slightly higher results. This shows that, when using text corpora, the current state of the art method (learn embeddings independently and then map them) is more effective than our method to concatenate the corpus and use the dictionary as a constraint in the loss function to learn bilingual embeddings jointly. We will show, in the following sections, that the situation is reversed when dealing with knowledge bases.

Note that the results in Table \ref{tab:bic} on the cross-lingual datasets are lower but close to those reported on monolingual datasets. For instance, \citep{baroni-dinu-kruszewski:2014:P14-1} report results of 0.73 on WS and 0.83 on RG using skipgram, while the best results in Table \ref{tab:bic} are 0.69 and 0.81, respectively. Note that the numbers are not comparable, as the translations might increase the difficulty of the task\footnote{Note also that these figures were obtained using different corpora and hyperparameters, so they are only mentioned as a point of reference.}. When comparing accross language pairs, the results for some language pairs are lower, with ENEU reporting the best scores and ENIT and ENES similar values. The variability across languages can be explained by a number of factors, including different difficulty of the datasets because of polysemy in the translations. A study of those factors is out of the scope of this article. Our work focuses on comparing the performance of different systems, which is fine as long as their relative performance is coherent across language pairs and datasets.

\subsection{Results using random walks on KBs (synthetic corpora)}
\label{sec:exp3}

In this set of experiments, we focus on the synthetic corpora produced using random walks over the wordnets, as described in Section \ref{sec:rw}, without access to text corpora. We  learn bilingual embeddings using the three methods tested in the previous section, but applied to the synthetic corpora, as follows: the state of the art \textsc{map}$_\text{kb}$, and our two methods \textsc{joint}$_\text{kb}$ and \textsc{jointc}$_\text{kb}$. In the case of \textsc{map}$_\text{kb}$, we learned monolingual embeddings from monolingual synthetic corpora. In order to have a comparable numbers of tokens (with respect to \textsc{map}$_\text{txt}$), we produced synthetic corpora of the same sizes as in Table \ref{tab:monotxt}. The mapping is then performed using the dictionary extracted from the bilingual wordnet, as in the previous section.
In the case of  \textsc{joint}$_\text{kb}$ and \textsc{jointc}$_\text{kb}$, we produced bilingual synthetic corpora totaling the same number of tokens as in Table \ref{tab:bitxt}. As mentioned in Section \ref{sec:exp}, the different sizes of the wordnets causes the bilingual random walk corpora to contain more words in one language than in the other. For instance, in ENES around 65\% of the tokens are in English, which raises to around 80\% in ENEU and ENIT. Regarding the other languages, in ESEU and ESIT around 65\% and 70\% of the tokens are in Spanish, respectively, and in EUIT Basque amounts to 60\% of the tokens. \textsc{jointc}$_\text{kb}$ uses the same bilingual constraints as \textsc{map}$_\text{kb}$.

\begin{table}[]
\centering
\begin{adjustbox}{width=1\textwidth}
\begin{tabular}{|l|l|ccc|ccc|r|l|}
\cline{3-8}
\multicolumn{1}{c}{\cellcolor[HTML]{FFFFFF}} & \multicolumn{1}{c|}{\cellcolor[HTML]{FFFFFF}} & WS            & \multicolumn{1}{l}{SL} & RG            & \multicolumn{1}{l}{WS} & SL            & \multicolumn{1}{l|}{RG} & \multicolumn{1}{l}{}       & \multicolumn{1}{l}{}                                             \\ \hline
                        
%{}                       & MORW                      & 5.0           & ---           & -6.1          & 3.1            & 1.5          & --- & {MORW}  & {}   \\ %\cline{2-9}                                                                                                        
%{}                       & \textsc{map}$_\text{txt}$ & \textbf{68.5}          & ---           & \textbf{81.5}          & 59.8           & \textbf{35.8}          & --- & {\textsc{map}$_\text{txt}$} & {}   \\ %\cline{2-9}                                       
{}                       & \textsc{map}$_\text{kb}$    & 65.1          & ---           & \textbf{88.2} & 55.8          & \textbf{47.2} & --- & {\textsc{map}$_\text{kb}$} & {}   \\ %\cline{2-9}                                                                     
{}                       & \textsc{joint}$_\text{kb}$  & \textbf{68.9} & ---           & 84.8          & 59.6          & 46.6          & --- & {\textsc{joint}$_\text{kb}$} & {}                       \\                                                        
{\multirow{-3}{*}{ENEU}} & \textsc{jointc}$_\text{kb}$ & 67.6          & ---           & 86.1          & \textbf{59.8} & 46.7          & --- & {\textsc{jointc}$_\text{kb}$} & {\multirow{-3}{*}{ENIT}}                       \\ \cline{1-10}                        
%{}                       & MORW                      & -5.9          & -1.5          & 8.3           & 0.4           & 0.9           & --- & {MORW}  & {}   \\ %\cline{2-9}                                                                                                         
%{}                       & \textsc{map}$_\text{txt}$ & \textbf{62.6} & \textbf{33.9} & \textbf{61.4} & \textbf{55.9} & \textbf{31.6} & --- & {\textsc{map}$_\text{txt}$} & {} \\ %\cline{2-9}                                                                    
{}                       & \textsc{map}$_\text{kb}$    & 62.9          & 46.1          & 77.1          & 44.9          & 38.5          & --- & {\textsc{map}$_\text{kb}$} & {}   \\ %\cline{2-9}                                                                     
{}                       & \textsc{joint}$_\text{kb}$  & 64.5          & 46.6          & 79.2          & \textbf{50.4} & \textbf{42.7} & --- & {\textsc{joint}$_\text{kb}$} & {}                       \\                                                        
{\multirow{-3}{*}{ENES}} & \textsc{jointc}$_\text{kb}$ & \textbf{64.7} & \textbf{48.2} & \textbf{79.7} & 49.1          & \textbf{42.7} & --- & {\textsc{jointc}$_\text{kb}$} & {\multirow{-3}{*}{ESIT}}                       \\ \cline{1-10}                        
%{}                       & MORW                      & 4.8           & ---           & -5.6          & 8.2           & ---           & --- & {MORW}  & {}   \\ %\cline{2-9}                                                                                                        
%{}                       & \textsc{map}$_\text{txt}$ & \textbf{64.4} & ---           & \textbf{65.1} & \textbf{60.9}  & ---           & --- & {\textsc{map}$_\text{txt}$} & {}   \\ %\cline{2-9}                                                                  
{}                       & \textsc{map}$_\text{kb}$    & 58.0          & ---           & \textbf{67.6} & \textbf{55.5} & ---           & --- & {\textsc{map}$_\text{kb}$}  & {}   \\ %\cline{2-9}                                                                    
{}                       & \textsc{joint}$_\text{kb}$  & 60.1          & ---           & 66.6          & 49.2          & ---           & --- & {\textsc{joint}$_\text{kb}$} & {}                       \\                                                        
{\multirow{-3}{*}{ESEU}} & \textsc{jointc}$_\text{kb}$ & \textbf{61.0} & ---           & 67.5          & 50.8          & ---           & --- & {\textsc{jointc}$_\text{kb}$} & {\multirow{-3}{*}{EUIT}}                       \\ \hline                             
\end{tabular}
\end{adjustbox}
\caption{Results (Spearman) on the cross-lingual datasets using synthetic, random-walk generated, corpora. \textsc{map}$_\text{kb}$ makes reference to results using separate monolingual random walk vector spaces plus the mapping. \textsc{joint}$_\text{kb}$ and \textsc{jointc}$_\text{kb}$ refer to the joint bilingual vector space, without and with bilingual constraints, respectively. Best result for each dataset and language pair in bold.}
\label{tab:birw}
\end{table}

Table \ref{tab:bic2} shows the performance on the six pairs of languages. The results for the three methods are mixed. In average, it seems all three methods do equally well, with each method winning in a comparable number of datasets. This means that, in the case of synthetic corpora, learning a joint space from the bilingual synthetic corpora  matches the results of the method that first learns from monolingual synthetic corpora and then learns the mapping. The lack of improvement for constraints might be caused by the fact that the synthetic corpora implicitly encodes translations equivalents, and thus the added bilingual constraints are redundant, in contrast to the previous section. 

%The  results when using the independently learned monolingual hybrid embeddings are again close to zero (MOHY). When applying the mapping to the hybrid embeddings (MOHY+MAP) the results improve, and in fact surpass those obtained with same method applied to genuine textual corpora of the same size (\textsc{map}$_\text{txt}$ row in Table \ref{tab:bic}) in all cases (languages and datasets) by a large margin, except for EUIT. This improvement is due to the information coming from the random walks over wordnet, and confirms the same effect observed in the monolingual case by \cite{goikoetxea2016single}. 

%Random walks alone (\textsc{joint}$_\text{kb}$) also provide strong results which are slightly improved with bilingual constraints (\textsc{jointc}$_\text{kb}$). Note that these embeddings only use the information in WordNet. The comparison with text-only embeddings (\textsc{map}$_\text{txt}$ row in Table \ref{tab:bic}) shows that they both perform conparably on WS, which is a relatedness dataset, while \textsc{jointc}$_\text{kb}$ clearly beats the text-only embeddings on RG and SL, which are genuine similarity datasets. This confirms that wordnets excel in capturing similarity information. 

A comparison to the results of the best text-only embeddings (\textsc{map}$_\text{txt}$ in Table \ref{tab:bic}, previous section) shows that the knowledge-based embeddings (all three methods in this section) perform consistently better on the similarity datasets by a large maring (RG and SL), while they underperform in the relatedness dataset (WS).

\subsection{Results using a combination of text and random walks (hybrid corpora)}
\label{sec:exp4}

In these set of experiments, we combine textual corpora and synthetic corpora (we will refer to these as \emph{hybrid} corpora, cf. end of Section \ref{sec:birw}), using the same methods as in the previous section. \textsc{joint}$_\text{hyb}$ and \textsc{jointc}$_\text{hyb}$ are applied to the concatenation of the monolingual textual corpora and synthetic corpora produced using bilingual random walks over the wordnets. \textsc{map}$_\text{hyb}$, on the other hand, learns the monolingual embeddings from two monolingual corpora which combine, each of them, monolingual text corpus and monolingual synthetic corpus. As in the previous sections, we use a dictionary extracted from the bilingual wordnet both for constraints and for learning the mappings. 

In the case of  \textsc{joint}$_\text{hyb}$ and \textsc{jointc}$_\text{hyb}$ we created hybrid bilingual corpora of the same size as the bilingual corpora in the previous sections (cf. Table \ref{tab:bitxt}) with a balanced number of tokens in the two languages, in order to control for corpus size and to show that the performance differences are not due to using larger corpora.  As mentioned in Section \ref{sec:exp}, the different sizes of the wordnets causes the random walk corpora to contain more words in one language than in the other. In preliminary experiments we saw that balancing the number of tokens coming from genuine textual corpora and random walks produced better results. We thus have produced hybrid corpora which fulfill the following constrains: they contain the same number of tokens as those in Table \ref{tab:bitxt}, with a balanced number of tokens in each language (that is 50\% each) and a balanced number of tokens between genuine and random walk corpora\footnote{We are aware that other distributions across languages and genuine/random walk corpora are possible, but let the exploration of other options for future work.}. The constraints  lead to the distribution of tokens shown in Table \ref{tab:lantok}.

Regarding \textsc{map}$_\text{hyb}$, monolingual embeddings were learnt from two monolingual hybrid corpora which combined textual and synthetic corpora, with sizes following the following restrictions: balanced number of tokens for both languages, and balanced number of tokens for each source of corpus (genuine and random walks). The total numbers of tokens correspond to those in Table \ref{tab:bitxt}.

\begin{table}[t]
\centering
\begin{tabular}{|c|c|rrr|rrr|c|c|}
\cline{3-8}
\multicolumn{1}{l}{}     & \multicolumn{1}{l}{}     & \multicolumn{1}{|c}{TXT}     & \multicolumn{1}{c}{RW} & \multicolumn{1}{c}{Total} & \multicolumn{1}{|c}{TXT} & \multicolumn{1}{c}{RW}      & \multicolumn{1}{c|}{Total} & \multicolumn{1}{l}{}    & \multicolumn{1}{l}{}                        \\ \hline
{}                       & EN & 32  & 128 &                       & 266 & 114 &                       & {ES} & {}   \\                                                
{\multirow{-2}{*}{ENEU}} & EU & 128 & 32  & \multirow{-2}{*}{320} & 114 & 266 & \multirow{-2}{*}{760} & {IT} & {\multirow{-2}{*}{ESIT}} \\\hline %\hhline{|========|} 
{}                       & ES & 53  & 107 &                       & 72  & 308 &                       & {EN} & {}                       \\                            
{\multirow{-2}{*}{ESEU}} & EU & 107 & 53  & \multirow{-2}{*}{320} & 308 & 72  & \multirow{-2}{*}{760} & {IT} & {\multirow{-2}{*}{ENIT}} \\\hline % \hhline{|========|}
{}                       & EU & 64  & 96  &                       & 147 & 283 &                       & {EN} & {}                       \\                            
{\multirow{-2}{*}{EUIT}} & IT & 96  & 64  & \multirow{-2}{*}{320} & 283 & 147 & \multirow{-2}{*}{860} & {ES} & {\multirow{-2}{*}{ENES}} \\ \hline                     
\end{tabular}
\caption{Amount of tokens (in millions) in the six bilingual hybrid corpora, split according to language and source of the tokens (genuine textual - TXT - and random walks - RW). }
\label{tab:lantok}
\end{table}

\begin{table}[]
\centering
\begin{adjustbox}{width=1\textwidth}
\begin{tabular}{|l|l|ccc|ccc|r|l|}
\cline{3-8}
\multicolumn{1}{c}{\cellcolor[HTML]{FFFFFF}} & \multicolumn{1}{c|}{\cellcolor[HTML]{FFFFFF}} & WS            & \multicolumn{1}{l}{SL} & RG            & \multicolumn{1}{l}{WS} & SL            & \multicolumn{1}{l|}{RG} & \multicolumn{1}{l}{}       & \multicolumn{1}{l}{}                                             \\ \hline
                        
%{}                       & MOHY     & -0.40         & ---           & 22.2          & 2.5           & 3.9           & --- & {MOHY} & {}                       \\  
{}                       & \textsc{map}$_\text{hyb}$ 		& 74.1         & ---           & 86.8          & 61.7          & 41.5          & --- & {\textsc{map}$_\text{hyb}$} & {}                       \\ %\cline{2-9}                       
%{}                       & \textsc{joint}$_\text{kb}$      & 68.9          & ---           & 84.8          & 59.6          & 46.6          & --- & {\textsc{joint}$_\text{kb}$} & {}                       \\                                    
%{}                       & \textsc{jointc}$_\text{kb}$		& 67.6          & ---           & 86.1          & 59.8          & 46.7          & --- & {\textsc{jointc}$_\text{kb}$} & {}                       \\                                    
{}                       & \textsc{joint}$_\text{hyb}$     & 74.0          & ---           & 85.6          & 63.5          & 46.6          & --- & {\textsc{joint}$_\text{hyb}$} & {}                       \\                                    
{\multirow{-3}{*}{ENEU}} & \textsc{jointc}$_\text{hyb}$ & \textbf{74.7} & ---           & \textbf{87.2} & \textbf{64.7}    & \textbf{47.6} & --- & {\textsc{jointc}$_\text{hyb}$} & {\multirow{-3}{*}{ENIT}} \\ \cline{1-10}%\hline%\hhline{|==========|}%\hline
%{}                       & MOHY     & 5.0           & 3.6           & -0.5          & -0.6          & -0.1          & --- & {MOHY} & {}                       \\
{}                       & \textsc{map}$_\text{hyb}$ 		& 70.2          & 46.5          & 84.2         & 56.2          & 35.5          & --- & {\textsc{map}$_\text{hyb}$} & {}                       \\ %\cline{2-9}                       
%{}                       & \textsc{joint}$_\text{kb}$     & 64.5          & 46.6          & 65.4          & 50.4          & 42.7          & --- & {\textsc{joint}$_\text{kb}$} & {}                       \\                                    
%{}                       & \textsc{jointc}$_\text{kb}$ & 64.7          & 48.2          & 65.0          & 49.1          & 42.7          & --- & {\textsc{jointc}$_\text{kb}$} & {}                       \\                                    
{}                       & \textsc{joint}$_\text{hyb}$     & 70.2          & 48.6           & 84.8          & 56.8          & 40.7          & --- & {\textsc{joint}$_\text{hyb}$} & {}                       \\                                    
{\multirow{-3}{*}{ENES}} & \textsc{jointc}$_\text{hyb}$ & \textbf{72.0} & \textbf{50.2} & \textbf{85.7} & \textbf{58.8}    & \textbf{43.3} & --- & {\textsc{jointc}$_\text{hyb}$} & {\multirow{-3}{*}{ESIT}} \\ \cline{1-10}%\hline%\hhline{|==========|}       
%{}                       & MOHY     & 7.4           		& ---           & 4.5           & -3.3          & ---           & --- & {MOHY} & {}                       \\
{}                       & \textsc{map}$_\text{hyb}$ 		& 69.7          & ---           & 71.8          & 55.5          & ---           & --- & {\textsc{map}$_\text{hyb}$} & {}                       \\ %\cline{2-9}                       
%{}                       & \textsc{joint}$_\text{kb}$      & 60.1          & ---           & 66.6          & 49.2          & ---           & --- & {\textsc{joint}$_\text{kb}$} & {}                       \\                                    
%{}                       & \textsc{jointc}$_\text{kb}$ 	& 61.0          & ---           & 67.5          & 50.8          & ---           & --- & {\textsc{jointc}$_\text{kb}$} & {}                       \\                                    
{}                       & \textsc{joint}$_\text{hyb}$      & 67.7          & ---           & 70.8          & 56.6          & ---           & --- & {\textsc{joint}$_\text{hyb}$} & {}                       \\                                    
{\multirow{-3}{*}{ESEU}} & \textsc{jointc}$_\text{hyb}$ 	& \textbf{70.0} & ---           & \textbf{71.9} & \textbf{58.1} & ---           & --- & {\textsc{jointc}$_\text{hyb}$} & {\multirow{-3}{*}{EUIT}} \\ \hline                             
\end{tabular}
\end{adjustbox}
\caption{Results (Spearman) on the cross-lingual datasets using hybrid corpora, combining natural and synthetic text.  \textsc{map}$_\text{hyb}$ makes reference to results using separate hybrid monolingual vector spaces plus the mapping. \textsc{joint}$_\text{hyb}$ and \textsc{jointc}$_\text{hyb}$ refer to the joint bilingual vector space, without and with bilingual constraints, respectively. Best results for each dataset and language in bold.}
\label{tab:bic2}
\end{table}

%Further, we enforce cross-lingual similarity applying BIC to \textsc{joint}$_\text{hyb}$, and, as in section \ref{sec:exp2}, we compare our method head to head to MAP. In this case, we feed the latter with vector spaces trained in separate hybrid monolingual corpora (MOHY) which follow the same criteria as before; that is, same contribution of tokens across languages (determined by size of minority language), and same contribution of tokens in text and WordNet pseudo-corpora within each language. 

Table \ref{tab:bic2} presents the results of the three methods on the combined, hybrid, corpora. In this case, it is the joint method with bilingual constraints that consistently outperforms the other two methods in all datasets, showing that the constraints are effective in this setting and that our joint method is more effective than the mapping method. 
%The results when using the independently learned monolingual hybrid embeddings are again close to zero (MOHY).
In fact, \textsc{jointc}$_\text{hyb}$ yields the best results among all methods and embeddings (cf. Tables \ref{tab:bic} and \ref{tab:birw}), showing that our method to combine text and knowledge bases is more effective to the state-of-the-art mapping method.

%Finally, the table  shows the results of the embeddings learned over the hybrid bilingual corpus (\textsc{joint}$_\text{hyb}$). When enforcing the bilingual constraints (\textsc{jointc}$_\text{hyb}$) we improve results further, consistently yielding the best results for all datasets and language pairs with respect to all competing systems, including mapping using dictionaries (\textsc{map}$_\text{hyb}$ row). All in all, the combination of random walks and bilingual constraints thus produces the best results, and shows that random walks are able to infuse some of the information in wordnet beyond that of bilingual constraints, outperforming the state of the art in bilingual embeddings produced using dictionaries.  

%When applying the mapping to the hybrid embeddings (\textsc{map}$_\text{hyb}$) the results improve, and in fact surpass those obtained with same method applied to genuine textual corpora of the same size (\textsc{map}$_\text{txt}$ row in Table \ref{tab:bic}) in all cases (languages and datasets) by a large margin, except for EUIT. This improvement is due to the information coming from the randow walks over wordnet, and confirms the same effect observed in the monolingual case by \cite{goikoetxea2016single}. 

\begin{table}[]
\centering
\begin{tabular}{l|cccc|}
\cline{2-5}
\multicolumn{1}{c|}{\cellcolor[HTML]{FFFFFF}} 			& all  			& rel  			& sim  			& win \\ \hline
\multicolumn{1}{|l|}{MAP$_\text{txt}$}        			& \underline{58.3} & \underline{62.0}	& \underline{54.6} & \underline{1}\\
\multicolumn{1}{|l|}{JOINT$_\text{txt}$}      			& 46.0 			& 53.0 			& 39.1 			& 0    \\
\multicolumn{1}{|l|}{JOINTC$_\text{txt}$}               & 52.2 			& 57.8 			& 46.6 			& 0    \\\hline
\multicolumn{1}{|l|}{MAP$_\text{kb}$}                   & 58.2 			& 55.6 			& 60.8 			& \underline{1}\\
\multicolumn{1}{|l|}{JOINT$_\text{kb}$}                 & 60.0 			& \underline{58.8}	& 61.1 			& 0    \\
\multicolumn{1}{|l|}{JOINTC$_\text{kb}$}                & \underline{60.3}	& \underline{58.8}	& \underline{61.8}	& 0    \\\hline
\multicolumn{1}{|l|}{MAP$_\text{hyb}$}                  & 62.8 			& 64.6 			& 61.1 			& 0    \\
\multicolumn{1}{|l|}{JOINT$_\text{hyb}$}                & 63.8 			& 64.8 			& 62.9 			& 0    \\
\multicolumn{1}{|l|}{JOINTC$_\text{hyb}$}               & \textbf{\underline{65.3}} & \textbf{\underline{66.4}} & \textbf{\underline{64.3}} & \textbf{\underline{10}}   \\ \hline
\end{tabular}
\caption{Summary of results: average across all datasets and languages (all), average on relatedness datasets (rel), average on similarity datasets (sim) and number of best results across all datasets and languages (win). Underline for best in each group, and bold for best overall. }
\label{tab:summ}

\end{table}

\begin{table}[]
\centering
\begin{tabular}{ll|ccc|}
\cline{3-5}
     &    				                         & WS 	       & SL          & RG  \\ \hline
\multicolumn{1}{|l|}{\multirow{2}{*}{ENES}} & NASARI$_\text{uni}$ &53.8         &44.0         &82.0        \\
\multicolumn{1}{|l|}{}                      & JOINTC$_\text{hyb}$ &\textbf{72.0}&\textbf{50.2}&\textbf{85.7}\\ \hline
\multicolumn{1}{|l|}{\multirow{2}{*}{ENIT}} & NASARI$_\text{uni}$ & 53.1         &\textbf{50.4}&--- \\
\multicolumn{1}{|l|}{}                      & JOINTC$_\text{hyb}$ & \textbf{64.7}& 47.6        &---\\\hline
\multicolumn{1}{|l|}{\multirow{2}{*}{ESIT}} & NASARI$_\text{uni}$ & 50.7&\textbf{48.8}& --- \\
\multicolumn{1}{|l|}{}                      & JOINTC$_\text{hyb}$ &  \textbf{58.8}&43.4&---\\\hline

\end{tabular}
\caption{Results (Spearman) on cross-lingual datasets for the state-of-the-art NASARI unified (\textsc{nasari}$_\text{uni}$) and our system. The average performance are, respectively, 54.7 and 60.3, with our method winning in five out of the seven datasets.}
\label{tab:nasaricmp}
\end{table}

\section{Summary of the main results and comparison to the state of the art}

Table \ref{tab:summ} summarizes the most relevant results with four figures: average performance across all datasets and languages (all, 12 datasets), average performance on relatedness datasets (rel, average of 6 datasets), average performance on similarity datasets (sim, 6 datasets),  and how many datasets does each method outperform the rest (win, 12 datasets). The summary shows clearly that our proposal to produce hybrid knowledge-based embeddings gets the best overall results, also in relatedness and similarity, and wins in 10 datasets out of the 12.

In addition, Table \ref{tab:nasaricmp} reports the results of the state-of-the-art NASARI which has been shown to outperform alternative methods in cross-lingual similarity \cite{camacho2016nasari}. We used the publicly available cross-lingual NASARI vectors that provided the best cross-lingual results\footnote{http://lcl.uniroma1.it/nasari/\#two} (NASARI$_\text{uni}$). They were readily available for English, Spanish and Italian, but the vectors for Basque were not available. We were generously assisted by one of the authors, so we could replicate their best results for English-Spanish and ensure that the rest of the results were correct. The table show that \textsc{jointc}$_\text{hyb}$ beats NASARI in all English-Spanish similarity and relatedness datasets, also in English-Italian and Spanish-Italian relatedness, but not on the similarity datasets. The average performances are, respectively, 60.3 and 54.7, with our method winning in five out of the seven datasets. This is despite our algorithm using just wordnets, in contrast to the much richer BabelNet. In fact, our lower results for Italian could be partially explained by the fact that the Italian wordnet is loosely aligned to the English WordNet, while BabelNet provides a tighter and richer integration. We think that our method combined with the richer BabelNet has the potential for further improvements.

%, which compares our best method (\textsc{jointc}$_\text{hyb}$) with the also state-of-the-art NASARI method in three of our language pairs. Note that NASARI embeddings have been extracted from BabelNet, which comprises several knowledge base sources (including Wikipedia and wordnet), whereas ours makes use of just wordnet. We used the publicly available cross-lingual NASARI embeddings\footnote{http://lcl.uniroma1.it/nasari/\#two} (NASARI$_\text{uni}$) for English, Spanish and Italian, thus rejecting from the comparison the three language pairs that included Basque. Along with the NASARI cross-lingual embeddings, in the ENES pair we have added two very competitive baselines which use English as pivot language: a wordnet-based (\textsc{adw}$_\text{pivot}$) method \citep{pilehvar2013align} and the best Word2vec model (\textsc{W2V}$_\text{pivot}$) in \citep{baroni2014don}.
%Even if the methods are not comparable

We will now summarize the main findings in this article.

\textbf{Random walks over multilingual wordnets improve results over just using dictionaries}: According to our experiments, the state of the art in bilingual embeddings using monolingual corpora and bilingual dictionaries (row  \textsc{map}$_\text{txt}$) underperforms our best system (\textsc{jointc}$_\text{hyb}$) in all datasets except one (WS for EUIT). Overall, we are able to improve results an average of $7$ points, with the biggest gain on similarity (10 points, cf. Table \ref{tab:summ}). This shows that multilingual wordnets are not mere bilingual dictionaries, and that our random walk method is able to exploit the internal structure of wordnets. %, which is key to improve results. 

\textbf{Multilingual wordnets are specially strong on similarity}: The information in the multilingual wordnets (without using textual corpora) is more effective than textual embeddings and bilingual dictionaries. Any of the methods to exploit random walks (\textsc{map}$_\text{kb}$, \textsc{joint}$_\text{kb}$ or \textsc{jointc}$_\text{kb}$) beats \textsc{map}$_\text{txt}$ in all similarity datasets, even if they do not exploit the information in textual corpora. The situation is reversed for word relatedness, where KB alone is slightly below, and in fact, the combination of wordnets and text is best for word relatedness.

\textbf{The improvements work across all wordnets}: The good results are notable for pairs with large wordnets like English and Spanish, but carry on to pairs with smaller wordnets like Basque, or loosely aligned wordnets like Italian. 
Note that the performance in ENIT is lower than in ENEU and ENES, either with our method or with MAP. This could be caused by the different structure of the Italian wordnet, and some information loss was to be expected because of the looser mapping. In the future, we would like to explore whether a different method to represent the graph for loosely aligned wordnets could be more effective. 

\textbf{Combining wordnets and textual corpora is very effective}: From another perspective, it is thanks to the combination of textual corpora and knowledge bases, properly exploited by our random walk and joint learning method with added bilingual constraints (\textsc{jointc}$_\text{hyb}$) that we overperform the state-of-the-art in text-based bilingual embeddings (\textsc{map}$_\text{txt}$ in Table \ref{tab:summ}). This fact is stressed by the fact that the joint method needs the synthetic corpora based on  random walks  to outperform \textsc{map}$_\text{txt}$. Regarding the state of the art in hybrid knowledge- and text-based methods, we show that our embeddings overperform NASARI in 5 out of 7 datasets, even though our embeddings rely only on wordnets, in contrast to the richer BabelNet. The other two datasets involve the Italian wordnet, where our results are weaker (see above). 
%On the other hand, the NASARI cross-lingual embeddings (\textsc{nasari}$_\text{uni}$) are comparable to \textsc{jointc}$_\text{hyb}$, as they are computed by combining the content in Wikipedia articles and BabelNet senses. Comparing the \textsc{nasari}$_\text{uni}$'s results (cf. Table \ref{tab:nasaricmp}) with the text-only methods, the former method beats \textsc{map}$_\text{txt}$ just in the similarity datasets. From this point of view, our method proves to be more robust than NASARI's, as \textsc{jointc}$_\text{hyb}$'s performance in both relatedness and similarity is better than in the text-only methods .\todo[color=green]{JOSU: neuk sartu dut azkenengo bi esaldi hauek}

\textbf{The two-step mapping method and our joint methods produce good results}: All three methods, when applied to the hybrid corpora containing natural and synthetic text (\textsc{map}$_\text{hyb}$, \textsc{joint}$_\text{hyb}$ and \textsc{jointc}$_\text{hyb}$), are able to overperform the current state of the art, with our joint method using bilingual constraints providing the best results. When using monolingual text alone (three top lines in Table \ref{tab:summ}), the mapping method uses the bilingual dictionary more effectively than our joint method with constraints. When using the synthetic corpora, the situation is reversed, with the joint method providing better results, specially when combined with the bilingual constraints in the loss function. This might be caused by the strong signal in the bilingual synthetic corpora used by the joint method, which the joint method does not take profit from.

\section{Conclusions and future work}
\label{sec:conc}

%\todo{REPLICABILITY: igual cross-lingual datasetak eta euskarazkoak distribuitu beharko genituzke, embedding hoberenekin batera (?)}

Bilingual embeddings are typically produced using monolingual corpora to learn the embeddings and a bilingual dictionary to learn a mapping from one embedding space to the other.
In this article we have proposed a novel method to produce a bilingual embedding space, based on monolingual corpora and bilingual wordnets. Bilingual wordnets can be used to produce bilingual dictionaries, but we propose a more effective method based on random walks that, in addition, extracts structural information. In fact we outperform the state of the art in bilingual embeddings \citep{artetxe-labaka-agirre:2016:EMNLP2016}, showing that we are  able to make effective use of the structure of bilingual wordnets. Our approach is based on a random walk algorithm over bilingual wordnets which emits lexicalizations in two languages as it traverses the wordnets. The bilingual corpus produced in this manner is combined with monolingual corpora which is then fed into skipgram, yielding bilingual embeddings. Further improvements are obtained incorporating bilingual constraints extracted from the wordnets into the skipgram loss function. 

From another perspective, we show that bilingual wordnets are more effective than bilingual dictionaries when producing bilingual embeddings. Given the increasing number of wordnets aligned with the English wordnet, and the fact that they can be used for any pair of languages which have a wordnet, our method shows great promise. For instance, larger bilingual knowledge bases like DBpedia or BabelNet could further augment the coverage of languages and further improve results. In fact, our method provides stronger results than NASARI \cite{camacho2016nasari} using just wordnets instead of the richer BabelNet, and it could be thus possible to improve further the results using richer resources. Domain-specific knowledge bases could be also useful when working on specific application areas. Note that, contrary to bilingual dictionaries and parallel corpora, once a wordnet is connected to the bilingual wordnets, it allows to produce bilingual embeddings of all other languages in the bilingual wordnet grid.

%Further, bilingual random walks combined with bilingual constraints have proved to be more effective for exploiting structural information from knowledge bases than the much more complex techniques like NASARI. Given the effectivity of our method, feeding it with complementary knowledge base information (e.g. Wikipedia and wordnet) can be a very interesting alternative to using larger knowledge bases as mentioned in the previous paragraph.\todo[color=green]{JOSU:neuk sartutakoa}

Our research opens a new avenue for producing multilingual embedding spaces, as it can work with larger Knowledge Bases like DBpedia or BabelNet, and can be easily extended from bilingual to multilingual embeddings. In addition to applying the method to other knowledge bases, we plan to explore further the different parameters of corpus combination. Being a first article using random walks for producing bilingual embeddings, we think there is still room for improvement. For instance, we would like to exploit multilingual wordnets and check whether multilingual embeddings for more than two languages are as effective as bilingual embeddings. 

The software and embedding sets that performed best, as well as the cross lingual
evaluation datasets and evaluation scripts are publicly available, with
instructions to obtain out-of-the-box replicability\footnote{\emph{url to be
    included upon acceptance}}. All software and resources of our method are open source.

\section{Acknowledgements}
\label{sec:ack}
We  thank  Mikel Artetxe and Jose Camacho-Collados  for  helping to run their systems, and Iraide Zipitria, Larraitz Uria and Izaskun Aldezabal for their contributions when building the Basque similarity datasets. This research was partially supported by a Google Faculty Research Award, the Spanish MINECO (TUNER TIN2015-65308-C5-1-R and MUSTER PCIN-2015-226) and the UPV/EHU (excellence research group). Josu Goikoetxea enjoys a grant from the University of the Basque Country. 

\clearpage
%\section*{References}
\bibliography{mybibfile}

\end{document}